# Exploring and Testing Skill-Based Behavioral Profile Annotation: Human Operability and LLM Feasibility under Schema-Guided Execution

Yufeng WU

City University of Hong Kong

## Abstract

Behavioral Profile (BP) annotation is difficult to automate because it requires simultaneous coding across multiple linguistic dimensions. We treat BP annotation as a bundle of annotation skills rather than a single task and evaluate LLM-assisted BP annotation from this perspective. Using 3,134 concordance lines of 30 Chinese metaphorical color-term derivatives and a 14-feature BP schema, we implement a skill-file-driven pipeline in which each feature is externally defined through schema files, decision rules, and examples. Two human annotators completed a two-round schema-only protocol on a 300-instance validation subset, enabling BP skills to be classified as directly operable, recoverable under focused re-annotation, or structurally underspecified. GPT-5.4 and three locally deployable open-source models were then evaluated under the same setup.

Results show that BP annotation is highly heterogeneous at the skill level: 5 skills are directly operable, 4 are recoverable after focused re-annotation, and 5 remain structurally underspecified. GPT-5.4 executes the retained skills with substantial reliability (accuracy = 0.678, $\kappa$ = 0.665, weighted F1 = 0.695), but this feasibility is selective rather than global. Human and GPT difficulty profiles are strongly aligned at the skill level (r = 0.881), but not at the instance level (r = 0.016) or lexical-item level (r = -0.142), a pattern we describe as shared taxonomy, independent execution. Pairwise agreement further suggests that GPT is better understood as an independent third skill voice than as a direct human substitute. Open-source failures are concentrated in schema-to-skill execution problems. These findings suggest that automatic annotation should be evaluated in terms of skill feasibility rather than task-level automation.

## 1. Introduction

Behavioral Profile (BP) annotation is one of the most demanding forms of corpus-based lexical analysis because it requires a single concordance line to be coded simultaneously across multiple linguistic dimensions. In a typical BP workflow, an annotator does not make one decision per instance, but many: lexical, syntactic, semantic, collocational, and pragmatic judgments must all be assigned within the same schema. This multidimensionality gives BP analysis much of its explanatory power, but it also makes large-scale annotation costly, difficult to standardize, and difficult to reproduce.

Recent work on large language models (LLMs) has renewed interest in the possibility of automating linguistic annotation. In many single-feature tasks, frontier LLMs have been shown to approach, and in some settings match, non-expert human annotators. BP annotation, however, poses a different challenge. It is not a single classification task repeated many times, but a bundle of heterogeneous annotation operations that vary substantially in how easily they can be operationalized from a schema alone. Treating BP annotation as a monolithic task therefore risks obscuring the more relevant methodological question: not whether "BP annotation" can be automated in general, but which parts of it can.

This paper argues that BP annotation is better understood as a bundle of annotation skills. Each BP feature corresponds to an underlying annotation operation with its own label inventory, decision rules, and evidential requirements. Some skills depend primarily on surface-observable cues, such as local grammatical position or overt co-occurrence markers. Others require broader interpretation, ontological categorization, or discourse-sensitive judgment. Once BP annotation is decomposed in this way, the relevant automation question becomes skill-specific: which annotation skills are executable under schema-guided conditions, and which remain unstable even for human annotators?

Our approach is not only conceptual but operational. The annotation pipeline examined in this study is skill-file-driven rather than prompt-only. BP skills are externally defined in schema files that specify labels, decision rules, and examples; prompt templates inject these skill definitions into model inputs; and a runner configuration binds schema resources, prompts, decoding defaults, and expected input structure into a reusable execution system. The result is an annotation workflow that is externally structured, auditable, and repeatable, rather than dependent on ad hoc prompt phrasing. In this sense, the present study evaluates not bare LLM performance, but LLM performance under explicit skill-grounded execution conditions.

Using this framework, we analyze 14 BP features applied to 3,134 concordance lines of 30 Chinese metaphorical color-term derivatives. Two human annotators labeled a validation subset in two independent schema-only rounds, allowing us to distinguish directly operable skills, skills recoverable under focused re-annotation, and structurally underspecified skills. GPT-5.4 and three locally deployable open-source models were then evaluated against the resulting gold standard. The results reveal a sharply stratified picture. At the skill level, human difficulty and GPT difficulty are strongly aligned ($r = 0.881$), but this shared structure collapses at the instance level ($r = 0.016$) and lexical-item level ($r = -0.142$). GPT reaches substantial agreement on the retained skills (accuracy = 0.678, $\kappa = 0.665$), while the open-source models fail in more localized but diagnostically revealing ways, especially in schema-to-label execution.

These findings motivate a different interpretation of LLM-assisted annotation. GPT-5.4 is not best understood as a direct human substitute. Instead, because it shares the same broad skill-difficulty topology as human annotators while exhibiting largely independent execution patterns at finer granularities, it is better interpreted as an independent additional annotation voice within a skill-screened workflow. Under this view, the central problem is no longer whether LLMs can replace annotators wholesale, but which skills can be reliably delegated, which require focused human recovery, and which remain too structurally underspecified for schema-only automation.

The contribution of this paper is twofold. Empirically, it shows that skill feasibility in BP annotation is uneven and strongly conditioned by schema-level operability. Methodologically, it proposes a skill-based view of annotation design in which pilot screening, focused re-annotation, and LLM deployment are organized around the executability of individual skills rather than the success or failure of an entire annotation task. On this view, automatic annotation is feasible only for the subset of BP skills that are operationalizable under schema-guided execution.

Against this background, the study addresses three related questions about the structure, operability, and automation of BP annotation under schema-guided conditions:

(1) Can Behavioral Profile annotation be productively decomposed into distinct annotation skills, and how operable are these skills under schema-only human annotation conditions?

(2) To what extent is annotation difficulty shared between humans and GPT-5.4 across different levels of analysis, and what does this imply about GPT-5.4's role in BP annotation workflows?

(3) What kind of skill-based workflow is most appropriate for semi-automatic BP annotation under schema-guided conditions?

## 2. Related Work

The skill-based perspective developed in this paper draws on three strands of prior work that have so far been pursued largely in isolation. The first is the Behavioral Profile (BP) tradition in corpus-based cognitive semantics, which provides the multi-dimensional annotation schema under study. The second is the rapidly growing literature on LLM-assisted annotation, which has largely evaluated models at the task level rather than at the level of individual annotation operations. The third is work treating human disagreement as information about task structure rather than as noise, which reframes what annotation reliability should be expected to measure. Section 2.1 reviews BP annotation and motivates the move from "one task, many fields" to "many skills, one schema." Section 2.2 locates the paper within LLM-assisted annotation and introduces the shift from task automation to skill feasibility. Section 2.3 connects human disagreement to the notion of shared skill difficulty. Section 2.4 synthesizes these strands into the skill-based framework that the rest of the paper operationalizes.

### 2.1 Behavioral Profile annotation as a multi-skill annotation problem

Behavioral Profile (BP) analysis has become one of the most influential corpus-based approaches to lexical semantics because it captures lexical behavior across multiple linguistic dimensions simultaneously. Rather than reducing a word to one distributional cue or one semantic variable, BP analysis encodes each concordance line for a set of lexical, syntactic, semantic, collocational, and pragmatic properties, thereby producing an instance-by-feature matrix that can be analyzed quantitatively. This multi-level design is precisely what gives BP its explanatory power in studies of near-synonymy, polysemy, semantic change, and distributional differentiation (Divjak & Gries, 2009; Gries, 2010).

At the same time, the BP tradition has generally treated the annotation schema as a unified coding instrument rather than as a bundle of distinct annotation operations. In practice, however, BP annotation is heterogeneous. Identifying word class, deciding construction type, assigning semantic role, and evaluating semantic prosody are not merely multiple labels within one task; they require different forms of evidence, different inferential depth, and different kinds of operational criteria. Some depend mainly on local structural cues, whereas others require interpretation across broader contextual or conceptual dimensions. This heterogeneity is already implicit in BP methodology itself, even if it has not usually been formalized in terms of distinct executable skills (Divjak & Gries, 2009; Gries, 2010). The present study therefore departs from the default "single task, multiple fields" view and instead treats each BP feature as an annotation skill unit. This reframing does not reject the BP tradition; rather, it makes explicit an operational distinction that has long been implicit in BP practice.

## 2.2 LLM-assisted annotation: from task automation to skill feasibility

Recent work on large language models has substantially expanded the methodological conversation around linguistic annotation. Across NLP and corpus linguistics, LLMs have been evaluated on tasks such as text classification, pragmatic annotation, grammatical categorization, and human-alignment-style labeling, often with results suggesting that frontier models can approach or in some settings even match non-expert human annotators under carefully designed prompting conditions (Movva et al., 2024; Törnberg, 2024; Fang et al., 2025). Within corpus-oriented work more specifically, a key methodological trend is the use of increasingly explicit task instructions, schema constraints, and prompt structuring rather than unconstrained free-form querying (Törnberg, 2024; Zhang et al., 2023).

However, most of this literature evaluates tasks at the level of the task as a whole. Even when multiple variables are annotated, the underlying logic is typically: define a task, run the model, compare its outputs to human labels, and report overall performance. This design is useful, but it leaves unresolved a deeper question that becomes unavoidable in multi-level annotation: when model performance varies sharply across parts of the schema, what is the relevant unit of explanation? Is the issue the model, the prompt, the training data, or the annotation operation itself? Recent work on annotation alignment and LLM-assisted subjective annotation has shown that model–human similarity can vary substantially by task and by judgment type, which makes this question especially salient in multi-dimensional annotation settings (Movva et al., 2024; Schroeder et al., 2025).

For BP annotation, this question is especially acute because the annotation schema already contains internally distinct operations. A model may perform very well on one BP feature and very poorly on another, even within the same instance and under the same prompt. In such cases, treating the schema as a single task can hide the fact that what is being automated is not “BP annotation” in general, but only a subset of BP annotation skills. The present study therefore extends the LLM-assisted annotation literature by shifting the analytic focus from task-level automation to skill feasibility: the extent to which a particular annotation operation, as defined in a schema, can be executed reliably under schema-guided conditions. This move is also consistent with work on schema-guided prompting, which treats task definitions as structured resources rather than mere prompt wording (Zhang et al., 2023).

## 2.3 From human disagreement to annotation difficulty as a task property

A second body of relevant work concerns the status of disagreement in annotation. Recent research has argued that disagreement should not always be treated as mere annotator noise. In many linguistic tasks, human disagreement reflects genuine task difficulty, underspecification in the schema, or competing but defensible interpretations. This line of work has encouraged researchers to view disagreement as information about the task rather than as a nuisance to be eliminated. The implications are especially important for annotation tasks involving inference, evaluative judgment, or category boundaries that are not fully recoverable from surface form alone (Aroyo & Welty, 2015; Plank, 2022).

The present paper adopts this perspective but pushes it in a new direction. If human disagreement can be interpreted as a task property, then LLM difficulty should not be analyzed solely in terms of whether the model matches a gold standard. It should also be analyzed in relation to where the schema itself generates stable versus unstable decisions. This motivates the idea of shared skill difficulty: if both humans and LLMs find the same BP features difficult, then at least part of the difficulty lies in the operational structure of the skill rather than in the annotator alone. This perspective is compatible with recent work emphasizing that label variation is not just noise to be suppressed, but a phenomenon that reshapes data construction, modeling, and evaluation (Plank, 2022).

At the same time, agreement at one level does not imply agreement at another. A model can share the same broad difficulty landscape as humans while still making different mistakes on specific instances. This possibility has rarely been examined explicitly in prior LLM-assisted annotation work, which tends to summarize model performance with a single accuracy or agreement score. The present study therefore argues that annotation difficulty should be examined at multiple levels of granularity and that the relationship between human and LLM performance may change depending on whether one looks at features, instances, or lexical items. In this respect, the paper is closer to recent annotation-alignment work that treats model–human similarity as structured and partial rather than absolute (Movva et al., 2024).

### 2.4 Why a skill-based perspective is needed

Taken together, the limitations of prior work suggest the need for a more fine-grained framework. BP annotation provides an ideal test case because it is simultaneously

(1) multi-level,

(2) schema-dependent,

(3) labor-intensive, and

(4) internally heterogeneous in difficulty. These properties make it unsuitable for a purely monolithic notion of automation. Instead, what is needed is a framework that can distinguish among at least three possibilities:

directly operable skills, which can be executed reliably under schema-only conditions;

recoverable skills, which initially appear unreliable but become stable under focused re-annotation;

structurally underspecified skills, which remain unstable even after renewed attention and therefore cannot be safely automated under the current schema.

This proposal builds directly on BP methodology while also drawing on recent work that treats annotation as a structured interaction among schema design, operational constraints, and human label variation (Divjak & Gries, 2009; Gries, 2010; Plank, 2022).

A skill-based perspective also aligns naturally with the architecture of the annotation system used in this study. The pipeline is not based on free-form prompting alone, but on externally defined schema files, prompt templates, and execution configurations that together specify how each BP feature is to be treated. In other words, the annotation process is already operationalized as a set of externally defined, reusable, and auditable skill units. This view is compatible with schema-guided prompting research, which shows that externally structured task definitions can materially shape model behavior (Zhang et al., 2023).

This perspective changes the interpretation of model performance. Under a skill-based view, the question is not whether GPT-5.4 or an open-source model can "do BP annotation," but whether a given model can execute a given BP skill under externally structured conditions. This distinction is crucial for interpreting both success and failure. Strong performance on a subset of skills should be understood as evidence for partial, structured automation rather than wholesale task replacement. Conversely, open-source failure may reflect not a complete lack of linguistic competence, but a more localized breakdown in schema-to-skill execution, as seen in feature-concentrated off-schema outputs. Similar concerns about partial alignment and non-uniform model behavior have also been raised in recent LLM annotation studies (Movva et al., 2024; Schroeder et al., 2025).

For these reasons, this paper treats skill as the central analytical unit. The contribution is not to replace existing BP methodology or the broader LLM annotation literature, but to integrate them under a framework that better captures what multi-level annotation actually requires: not one model solving one task, but one system executing many distinct annotation skills with different operational properties.

## 3. Method

The study combines a corpus of Chinese metaphorical color-term derivatives, a 14-feature BP schema treated as a skill inventory, a two-round schema-only human annotation protocol, and a skill-file-driven model execution pipeline. Section 3.1 describes the corpus and the four aligned data sources used in the human–LLM comparison. Section 3.2 sets out the skill inventory and introduces the feature/skill distinction that the rest of the paper relies on. Section 3.3 specifies the two-round human annotation protocol and the two gold definitions derived from it. Section 3.4 describes the skill-file-driven execution of GPT-5.4 and the three open-source models. Section 3.5 lays out the three-level evaluation framework that organizes the empirical analyses in Section 4.

### 3.1 Data

The study is based on a corpus of 3,134 concordance lines involving 30 metaphorical derivatives of six Chinese basic color terms. The lexical inventory covers five derivatives for each of the six colors—black, white, red, yellow, blue, and green—yielding a balanced set of target lexemes distributed across multiple semantic domains. The 30-item inventory includes, for example, *hēichē* 'unlicensed taxi', *báilǐng* 'white-collar worker', *hóngbāo* 'red envelope', *huángle* 'fell through', *lánchóu* 'blue chip', and *lǜkǎ* 'green card'. Each target lexeme is represented by approximately one hundred concordance lines in the full dataset (mean = 104.47 lines per lexeme, range = 98–112, SD = 2.39), and the human-validated subset contains 300 instances, corresponding to 10 instances per target lexeme.

For the human–LLM comparison, four aligned data sources were used. First, a two-round human annotation file contains 4,200 instance-feature cells (300 instances × 14 features), including first-round labels, second-round re-annotations for initially disagreed cells, and final labels for both annotators. Second, a GPT-5.4 output file contains predictions for the full 3,134-instance dataset across the BP feature inventory. Although F7 was re-annotated in an updated version of the GPT output file, it does not enter the main GPT-facing analyses under the current evaluation setup because a large proportion of F7 cases involve non-predicative uses for which valency is not meaningfully assessable. Third, a merged open-source output file contains the annotations of three locally deployable models—Qwen2.5-7B, GLM-4-9B, and Yi-1.5-9B. The effective row count of this file falls short of the nominal ceiling (300 instances × 10 retained features × 3 models) because some cells are empty or off-schema; per-model and per-feature rates of null and off-schema output are reported in Table 8, and only rows with valid in-schema labels enter the agreement analyses in Section 4.6. Fourth, a target-word mapping file links the 300 human-validated instance IDs to their corresponding target lexemes and source color categories. All files were scanned, normalized, and aligned prior to analysis. Path readability checks and data quality checks confirmed that the required data and system files were available and internally consistent.

As in the preceding BP-oriented versions of this project, all label values were normalized before analysis in order to collapse numerically equivalent but string-distinct forms (e.g. 3, 3.0, and "3"). GPT feature column names were stripped of incidental whitespace, and all agreement calculations used a safe kappa function that returned missing values for degenerate single-label cases. Two gold definitions were retained for robustness analysis: a Round-1 gold, based only on cells on which the two annotators agreed in their first pass, and a final gold, based on cells on which the two annotators converged after the two-round schema-only procedure. The primary analyses in this paper use the final gold.

### 3.2 Skill inventory and schema design

The central methodological move in this study is to treat BP annotation not as a single task but as a bundle of annotation skills. Concretely, each BP feature is interpreted as a skill unit: an externally defined annotation operation with its own label inventory, decision rules, examples, and output constraints. Under this view, the BP schema is not merely a coding sheet but a structured repository of executable annotation skills.

The schema defines 14 BP features spanning five linguistic levels: lexical, syntactic, semantic, collocational, and pragmatic. These include, among others, word class, lexicalization degree, syntactic function, construction type, semantic role, collocational patterning, semantic prosody, and register. For the purposes of the present paper, each of these features is treated as one skill unit in the skill inventory. The inventory was reconstructed directly from the external schema file, which specifies for each feature:
(1) a unique feature identifier,
(2) the label set,
(3) explicit decision rules, and
(4) illustrative examples.

This design supports a distinction between feature and skill. A feature is the observable coded dimension that appears in the annotation table, whereas a skill is the underlying annotation operation required to assign that feature reliably. The distinction matters because different features may require qualitatively different kinds of skill execution. For example, assigning word class often depends on relatively local structural cues, while assigning semantic prosody or semantic role may depend on broader contextual interpretation. Treating all features as if they were instances of one homogeneous task would therefore conceal the heterogeneity that is central to BP annotation. The skill framing introduced here makes that heterogeneity explicit. To fix the usage convention for the remainder of the paper: *feature* refers to the observable coded dimension in the BP coding matrix (the column in Table 1), and *skill* refers to the externally specified annotation operation required to assign a value on that dimension. There is a one-to-one correspondence between features and skills in the present inventory, but the two terms are not synonyms: where we refer to what is being coded, we use *feature*; where we refer to the operation doing the coding, we use *skill*. Operability claims (directly operable, recoverable, structurally underspecified) are claims about skills, not about features.

The schema was implemented as part of a skill-file-driven annotation system. According to the system summary, the external schema file defines each skill unit together with its labels, decision rules, and examples; the prompt template injects these schema-derived definitions into the model input and enforces constrained JSON output; the runner configuration binds schema, prompts, decoding defaults, and expected input columns; and the main annotation script supports both single-skill execution (--feature F1) and full-inventory execution (--feature all), with resume checkpoints for interrupted runs. The annotation process is therefore externally structured, auditable, and reusable, rather than dependent on informal prompt engineering alone.

This table summarizes the main external resources used in the skill-file-driven annotation pipeline. Rather than relying on ad hoc prompt phrasing alone, the system externalizes feature-level annotation skills through schema files, prompt templates, execution configurations, and auxiliary control files. Together, these resources make the annotation workflow reusable, auditable, and reproducible.

Table 1. Skill inventory for BP annotation

| skill_id | skill_name | linguistic_level | number_of_labels | number_of_examples |
|---|---|---|---|---|
| F1 | Word Class & POS | lexical | 7 | 3 |
| F10a | Left Collocation | collocation | 8 | 3 |
| F10b | Right Collocation | collocation | 7 | 3 |
| F11 | Marker Co-occurrence | collocation | 8 | 3 |
| F12a | Semantic Prosody | pragmatic | 4 | 3 |
| F12b | Register | pragmatic | 5 | 3 |
| F2 | Lexicalization Degree | lexical | 5 | 3 |
| F3 | Word Formation | lexical | 6 | 2 |
| F4 | Syntactic Function | syntactic | 8 | 2 |
| F5 | Construction Type | syntactic | 10 | 2 |
| F6 | Syntactic Position | syntactic | 6 | 2 |
| F7 | Valency | semantic | 5 | 2 |
| F8 | Core Argument Type | semantic | 10 | 2 |
| F9 | Semantic Role | semantic | 8 | 2 |

### 3.3 Two-round schema-only human annotation

To assess the operability of BP skills under human execution, we adopted a two-round schema-only annotation protocol. Two annotators independently labeled a validation subset of 300 instances across the 14 BP features. Importantly, the annotators worked under schema-only conditions: they were given the same externally defined annotation schema, but no additional calibration session, no extra feedback loop, and no collaborative discussion during the annotation process. The design therefore operationalizes what the skill inventory itself makes possible without external adjudication.

Round 1 provides a measure of cold-take skill operability. It captures whether a skill can be executed reliably when multiple skills are being applied simultaneously to the same instance. Because BP annotation involves many parallel coding decisions, first-round performance reflects not only structural clarity but also the cognitive burden of multi-skill annotation. In Round 2, only cells that were unresolved in Round 1 were revisited independently by the two annotators. This produces a more focused setting in which the relevant skill is re-executed without the full first-pass load. The comparison between Round 1 and Round 2 therefore allows us to distinguish two sources of failure:

(1) cognitive-load difficulty, where disagreement is largely recoverable under focused re-annotation, and
(2) structural difficulty, where disagreement persists even after renewed attention.

This distinction is crucial for the present paper. Under a skill-based interpretation, the human annotation protocol is not used only to create a gold standard. It is also used to classify skills according to their operability profile. In the final summary of the skill-framed results, five skills were directly operable in Round 1, four additional skills became operable only after focused re-annotation, and five remained structurally underspecified under the schema-only regime. This makes the two-round protocol an integral part of the substantive analysis rather than a mere preprocessing step.

For the formal analyses, gold labels were derived in two ways. The Round-1 gold retained only cells where the two annotators agreed in their first-round independent coding. The final gold retained cells where the two annotators agreed either immediately or after the second-round re-annotation. The final gold serves as the main evaluation reference because it better reflects skill operability after controlling for first-pass cognitive load, while the Round-1 gold is used for robustness and comparison.

### 3.4 Skill-file-driven model execution

The model-based part of the study compares one frontier closed-source model and three locally deployable open-source models under the same skill-grounded execution framework. GPT-5.4 serves as the frontier closed-source model, while Qwen2.5-7B, GLM-4-9B, and Yi-1.5-9B represent the open-source baseline group. The GPT output file contains predictions for the full 3,134-instance dataset; the merged open-source file contains the corresponding outputs of the three local models. Throughout the paper, *GPT-5.4* refers to the OpenAI CHATGPT Thinking model accessed between 24 March 2026 and 2 April 2026; all predictions reported here were produced within that window under the same skill-file-driven pipeline. The three open-source models were run locally over the same period using their publicly released weights. Results may not reproduce exactly under later model snapshots even with identical prompts and schema files.

The key methodological point is that the models were not prompted in an unconstrained, free-form way. Instead, model execution was mediated by the same skill files used to define the annotation system. The schema externally specified the label inventories, decision rules, and examples associated with each skill. The prompt template then injected these skill definitions into the model input, while the runner configuration specified decoding defaults and input-column expectations. The script controlling local execution supported feature-wise runs and all-feature runs, meaning that the system was explicitly designed to allow both single-skill and inventory-level annotation. This makes it appropriate to speak of the models as skill executors rather than generic text generators.

This design also enables a more precise diagnosis of model failure. If a model fails, the failure need not be interpreted as lack of linguistic competence in a broad sense. It may instead reflect a breakdown in schema-to-skill execution, such as mapping the wrong label inventory to the right underlying distinction. This distinction is especially important for the open-source models, whose failures are shown below to be concentrated in particular features and often attributable to off-schema or format-related outputs rather than to uniformly poor language understanding.

Finally, because the entire annotation pipeline is externally structured, the model execution process is reproducible. The main script, the schema file, the runner configuration, and the prompt template jointly define a pipeline that can be rerun from an empty environment once dependencies are installed. In this sense, the study evaluates not an isolated model call but a reusable annotation workflow.

Table 2. External resources in the skill-file-driven annotation pipeline

| **Evidence item** | **File** | **Evidence** |
|---|---|---|
| Schema specification | bp_schema.yaml | The schema file defines feature identifiers, label inventories, decision rules, and illustrative examples. |
| Prompt template | prompt_all_core.j2 | The prompt template incorporates schema-derived label tables and enforces fixed JSON output keys. |
| Runner configuration | runner_config.json | The configuration file integrates schema resources, prompt templates, decoding defaults, and expected input columns. |
| Annotation script | unified_annotator.py | The script supports feature-wise execution, full-inventory execution, and checkpoint-based resume. |

### 3.5 Three-level evaluation framework

The empirical core of the paper lies in a three-level evaluation framework that examines shared difficulty and execution behavior at different levels of granularity.

#### Skill level

At the first level, the unit of analysis is the skill itself, corresponding to one BP feature. Here the question is whether the global difficulty profile of skills is shared between human annotators and GPT. Human-side difficulty is indexed by pilot inter-annotator agreement, while GPT-side difficulty is indexed by its agreement with the human gold. This level captures what we call shared skill difficulty: whether both human and model annotators find the same skills globally easy or hard.

Instance level

At the second level, the unit of analysis is the concordance-line instance. Here the question is whether the specific instances that are difficult for humans are also the same instances on which GPT fails. This level tests whether shared difficulty at the skill level also implies shared execution failure at the item level. If not, then humans and GPT may agree on which skills are difficult overall while still making different errors on specific instances.

Lexical-item level

At the third level, the unit of analysis is the target lexeme. Here the question is whether target words that are globally difficult for human annotation are likewise difficult for GPT. This level captures broader lexical aggregation effects and helps determine whether shared difficulty is a property of skill type or of lexical inventory.

The three-level framework is central because it separates two claims that are often conflated in LLM-assisted annotation studies. A model may look “human-like” in the aggregate because it tracks the same difficult features, yet still behave differently from humans in how it distributes its errors across instances or lexical items. This is precisely the distinction captured by the formulation shared taxonomy, independent execution, which is tested quantitatively in the Results section.

In addition to the three-level analyses, the study includes:

- pairwise agreement analysis among the two human annotators, GPT, and the three open-source models;
- feature-wise skill feasibility analysis;
- open-source invalid-output diagnosis, with special attention to schema-to-label mapping failures;
- and a workflow-oriented synthesis that classifies skills into deployment tiers.

Together, these analyses support the main methodological claim of the paper: skill-based automatic annotation is feasible only for the subset of BP skills that are operationalizable under schema-guided execution.

## 4. Results

The results are presented in seven subsections that together build the skill-based account of LLM-assisted BP annotation. Section 4.1 establishes that BP annotation is empirically heterogeneous and is best modeled as a bundle of skills. Section 4.2 classifies the 14 skills into directly operable, recoverable, and structurally underspecified categories under the two-round human protocol, and discusses the two borderline cases, F2 and F7. Section 4.3 reports GPT-5.4's execution performance on the retained skills and makes explicit the composition of the retained set. Section 4.4 presents the three-level analysis that yields the shared-taxonomy, independent-execution pattern. Section 4.5 examines the pairwise agreement structure among the two humans, GPT, and the open-source models. Section 4.6 treats the open-source models as a boundary test for local skill execution and diagnoses their failure modes. Section 4.7 synthesizes the findings into a skill-based annotation workflow.

### 4.1 BP annotation is empirically heterogeneous and is best modeled as a bundle of skills

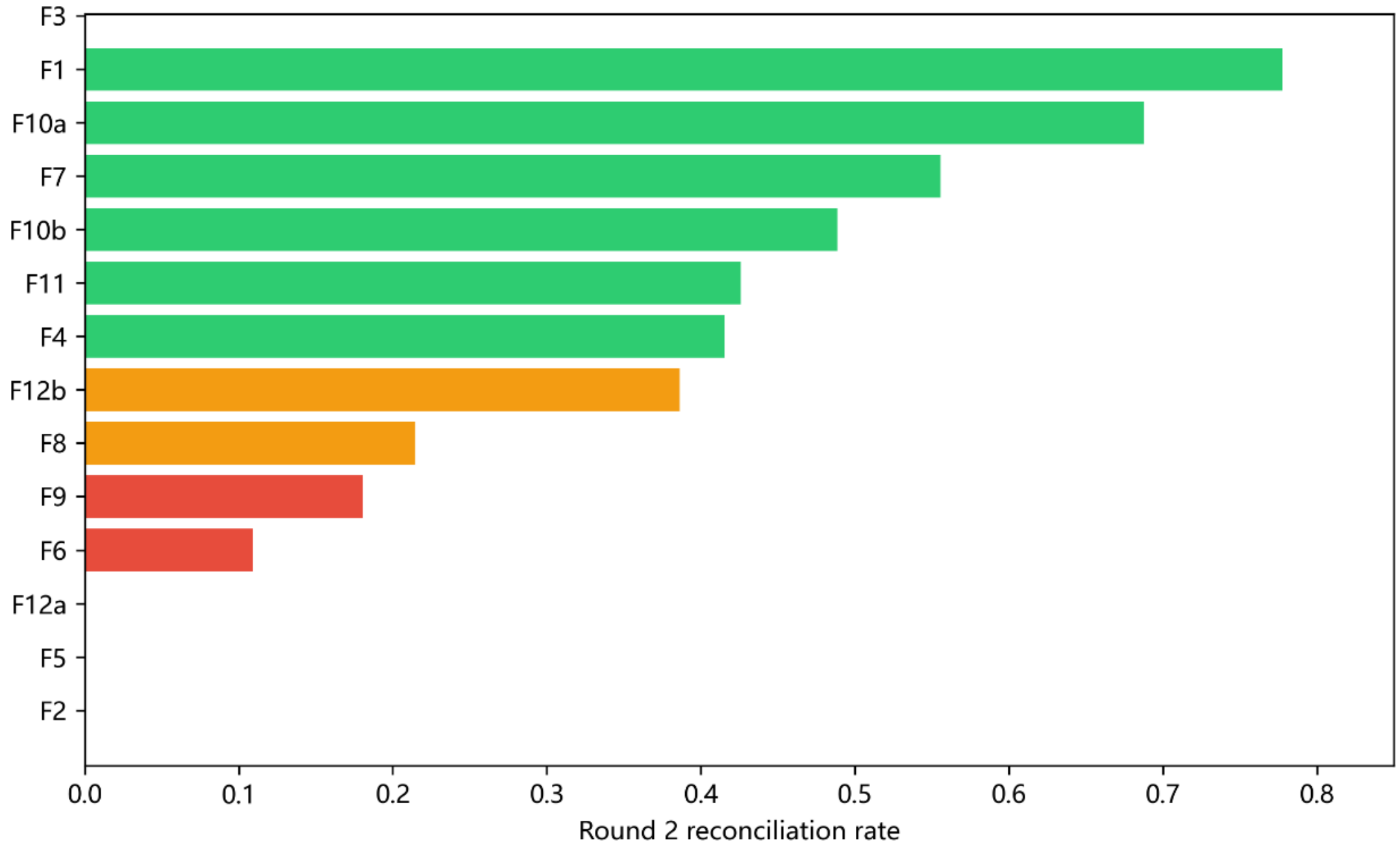

Figure 1. Round-2 reconciliation rates across BP skills

The first result of the study is that BP annotation cannot be meaningfully treated as a single homogeneous task. Once the 14 BP features are interpreted as 14 annotation skills, the internal heterogeneity of the task becomes immediately visible. Under schema-only conditions, the skills do not pattern uniformly: some are directly executable on a first pass, some become stable only after focused re-annotation, and others remain unstable even after a second independent attempt. In other words, the empirical evidence supports a bundle-of-skills view of BP annotation rather than a monolithic task view.

This heterogeneity is not just conceptual but operational. The annotation system itself is structured around feature-level skill units defined externally through schema files, prompt templates, and runner configurations. The skill-file-driven architecture therefore corresponds closely to the empirical structure of the task: the system executes what the analysis later shows to be genuinely distinct annotation operations. This alignment between system design and empirical task structure is important, because it means that the skill framing is not merely interpretive language applied after the fact; it reflects how the annotation workflow is actually organized.

### 4.2 Human skill operability: directly operable, recoverable, and structurally underspecified skills

Before presenting the results, we make the classification rule underlying skill_status in Table 3 explicit. Each BP skill is classified jointly on three quantities derived from the two-round schema-only protocol: the first-round κ (round1_kappa), the focused-re-annotation reconciliation rate (reconciliation_rate, defined as round2_reconciled / round1_disagreements), and the final-gold κ (final_kappa) computed on cells retained in the final gold. A skill is classified as directly operable when round1_kappa falls within at least the moderate range (operationalized here as ≥ .50) and final_kappa reaches the upper boundary of that range (≥ .60), indicating that the skill can be executed reliably on a first pass even under the full multi-skill annotation load. A skill is classified as recoverable under focused re-annotation when round1_kappa falls below the direct-operability threshold but final_kappa reaches at least .50 and the reconciliation rate is non-trivial (> .30), indicating that first-pass disagreement can be closed when annotators revisit the disputed cells in isolation. A skill is classified as structurally underspecified when either condition for recoverability fails: either final_kappa remains below .50, or the reconciliation rate is effectively zero so that a moderate final_kappa reflects only cells on which annotators happened to agree spontaneously in Round 1 rather than successful skill execution under focused attention. The classification therefore uses the reconciliation rate, not final_kappa alone, as the decisive evidence for whether a skill is operationalizable under the current schema. The κ cutoffs draw on the widely used agreement-interpretation scheme of Landis and Koch (1977), in which κ in the interval .41–.60 is typically interpreted as moderate agreement and κ in the interval .61–.80 as substantial agreement. On this basis, the Round-1 operability threshold of .50 is set within the moderate range, and the final-gold threshold of .60 at its upper boundary. By contrast, the reconciliation-rate cutoff of .30 is a pragmatic threshold adopted for the present study, reflecting the point at which roughly one-third of Round-1 disagreements can be resolved under focused re-annotation; to our knowledge, there is no standard threshold for this quantity in the annotation-reliability literature.

Table 3. Human skill operability under the two-round schema-only protocol

| feature | round1_kappa | final_kappa | gold_coverage | reconciliation_rate | round1_disagreements | round2_reconciled |
|---|---|---|---|---|---|---|
| F1 | 0.889 | 0.977 | 0.990 | 0.778 | 9 | 7 |
| F2 | 0.060 | 0.652 | 0.810 | 0.000 | 130 | 0 |
| F3 | | | 1.000 | | 0 | 0 |
| F4 | 0.612 | 0.781 | 0.853 | 0.416 | 77 | 32 |
| F5 | 0.036 | 0.172 | 0.283 | 0.000 | 239 | 0 |
| F6 | 0.297 | 0.390 | 0.510 | 0.109 | 165 | 18 |
| F7 | 0.211 | 0.828 | 0.063 | 0.556 | 9 | 5 |
| F8 | 0.393 | 0.558 | 0.660 | 0.215 | 135 | 29 |
| F9 | 0.508 | 0.626 | 0.710 | 0.181 | 72 | 13 |
| F10a | 0.106 | 0.693 | 0.753 | 0.688 | 237 | 163 |
| F10b | 0.382 | 0.698 | 0.777 | 0.489 | 131 | 64 |
| F11 | 0.199 | 0.560 | 0.663 | 0.426 | 176 | 75 |
| F12a | 0.530 | 0.726 | 0.827 | 0.000 | 86 | 0 |
| F12b | 0.522 | 0.711 | 0.797 | 0.386 | 101 | 39 |

The two-round schema-only human protocol reveals a three-way stratification of BP skills. First, five skills meet direct operability criteria in Round 1, indicating that they can be executed with acceptable reliability even when annotators are handling the full multi-skill burden of BP annotation. Second, four additional skills become reliable only after focused re-annotation. These are skills whose first-pass instability appears to reflect cognitive-load difficulty rather than irreducible ambiguity. Third, five skills remain structurally underspecified: even after renewed attention, they do not achieve stable execution under schema-only conditions.

Two entries in Table 3 deserve specific comment because their final-kappa values alone could be read as inconsistent with their classification. F2 (Lexicalization Degree) reaches a final-gold κ of .652, which on its own would suggest moderate agreement, but is nonetheless classified as structurally underspecified. The reason is that the skill's first-round κ is effectively zero (κ = .060), and none of the 130 first-round disagreements were closed under focused re-annotation (reconciliation rate = 0). The moderate final-gold value therefore does not reflect successful skill execution after focused attention; it reflects the subset of cells on which the two annotators happened to arrive at the same label spontaneously in Round 1. Given the decision rule set out above, where the reconciliation rate is decisive, F2 fails the operationalizability condition. F7 (Valency) exhibits the opposite pattern: a high final-gold κ of .828 but a gold coverage of only 6.3 percent of cells. This is not evidence of weak operability. F7 retains the original five-label schema (without an explicit not-applicable label) and is only interpretable in contexts where the target derivative is used predicatively in an argument-taking construction. In a large proportion of concordance lines, however, the target item appears in non-predicative uses, so the feature is not meaningfully assessable and the corresponding cells are left empty rather than entered into the gold. Within the small set of cells for which F7 is applicable, first-round disagreement is low in absolute terms (9 cases), the reconciliation rate is .556, and the resulting final-gold κ is high. F7 is therefore recoverable in the sense relevant to this paper: when the skill does apply, annotators converge under focused re-annotation. The low coverage reflects sparse applicability in the corpus, not sparse agreement.

This distinction is methodologically important. A low first-pass agreement score does not by itself imply that a skill is unsuitable for automation or unusable in BP analysis. Some apparently weak skills are recoverable once annotators revisit only the disputed cells. These cases suggest that first-round failure may arise from the burden of executing many skills simultaneously rather than from a defect in the skill definition itself. By contrast, skills that remain unreliable after focused re-annotation are stronger candidates for being fundamentally underspecified in the schema-only setting. The present results therefore support a distinction between cognitive-load difficulty and structural difficulty, and show that this distinction can only be detected through a two-round protocol.

From the standpoint of annotation design, this means that human pilot annotation should not be reduced to a single agreement threshold. What matters is not only whether a skill appears unstable at first pass, but whether that instability is recoverable. The two-round protocol thus serves both as a gold-construction method and as a diagnostic of skill operability.

### 4.3 GPT as a skill executor: substantial but selective skill feasibility

Table 4. GPT-5.4 execution performance on retained BP skills

| feature | n | accuracy | kappa | macro_f1 | weighted_f1 | random_baseline |
|---|---|---|---|---|---|---|
| F1 | 297 | 0.832 | 0.582 | 0.559 | 0.855 | 0.333 |
| F2 | 243 | 0.638 | 0.336 | 0.360 | 0.683 | 0.333 |
| F4 | 256 | 0.723 | 0.593 | 0.483 | 0.725 | 0.200 |
| F8 | 198 | 0.520 | 0.369 | 0.333 | 0.549 | 0.125 |
| F9 | 211 | 0.649 | 0.424 | 0.235 | 0.688 | 0.333 |
| F10a | 226 | 0.628 | 0.538 | 0.605 | 0.633 | 0.143 |
| F10b | 233 | 0.734 | 0.640 | 0.565 | 0.743 | 0.167 |
| F11 | 199 | 0.513 | 0.338 | 0.370 | 0.533 | 0.125 |
| F12a | 247 | 0.749 | 0.611 | 0.748 | 0.748 | 0.333 |
| F12b | 239 | 0.690 | 0.556 | 0.650 | 0.699 | 0.250 |
| OVERALL | 2349 | 0.678 | 0.665 | 0.488 | 0.695 | |

On the subset of retained ten skills, GPT achieves accuracy = 0.678, Cohen's κ = 0.665, weighted F1 = 0.695, and macro F1 = 0.488. These values indicate that GPT can execute a subset of BP skills with substantial reliability under schema-guided conditions, but they also show that performance is selective rather than uniform. The gap between weighted F1 and macro F1 suggests that GPT is more reliable on dominant labels than on low-frequency categories, which is consistent with a skill-feasibility interpretation rather than a claim of complete automation.

The composition of the "retained ten" set deserves to be stated explicitly, since it could otherwise appear to be a performance-driven selection. The ten skills in Table 4 are those meeting two operational conditions: (i) the skill enters the main GPT evaluation setup, which excludes F7 because a large proportion of its corpus instances involve non-predicative uses rather than argument-taking contexts (see Section 3.1); and (ii) final-gold coverage is large enough to compute non-degenerate agreement statistics, which excludes F3 (no label variance in the gold), F5 (final-gold $\kappa = .172$, coverage $= .283$), and F6 (final-gold $\kappa = .390$, coverage $= .510$). This retention criterion is operational, not performance-driven: it is not applied to the GPT predictions, and it is not conditioned on human operability status. In particular, F2 and F8 are included in Table 4 despite being classified as structurally underspecified in Table 3, because their final-gold coverage is sufficient for agreement computation. Their presence in Table 4 is what produces the low-feasibility tier visible in Table 5, and the overall numbers reported above therefore include, rather than exclude, the more difficult skills in the retained set.

The central implication is that GPT should be evaluated at the level of skill feasibility, not at the level of "BP annotation" in general. On the retained skills, GPT is good enough to participate meaningfully in the annotation workflow; on the excluded or structurally underspecified skills, no such conclusion can be drawn. This pattern reinforces the main methodological argument of the paper: automatic annotation is not an all-or-nothing property of the BP task as a whole, but a property of individual skills within the schema.

These results also show why a skill-based framing is more informative than an undifferentiated task-level framing. A single pooled score could be misread as saying that GPT either "can" or "cannot" do BP annotation. In fact, the pooled result only makes sense once it is understood as the aggregate performance of GPT on a screened subset of executable skills. The unit of explanation is therefore the skill, not the task.

Table 5. Skill feasibility tiers for BP annotation

| feature | kappa | accuracy | feasibility_tier |
|---|---|---|---|
| F10b | 0.640 | 0.734 | high |
| F12a | 0.611 | 0.749 | high |
| F4 | 0.593 | 0.723 | medium |
| F1 | 0.582 | 0.832 | medium |
| F12b | 0.556 | 0.690 | medium |
| F10a | 0.538 | 0.628 | medium |
| F9 | 0.424 | 0.649 | medium |
| F8 | 0.369 | 0.520 | low |
| F11 | 0.338 | 0.513 | low |
| F2 | 0.336 | 0.638 | low |

4.4 Shared skill difficulty, but independent execution at finer levels

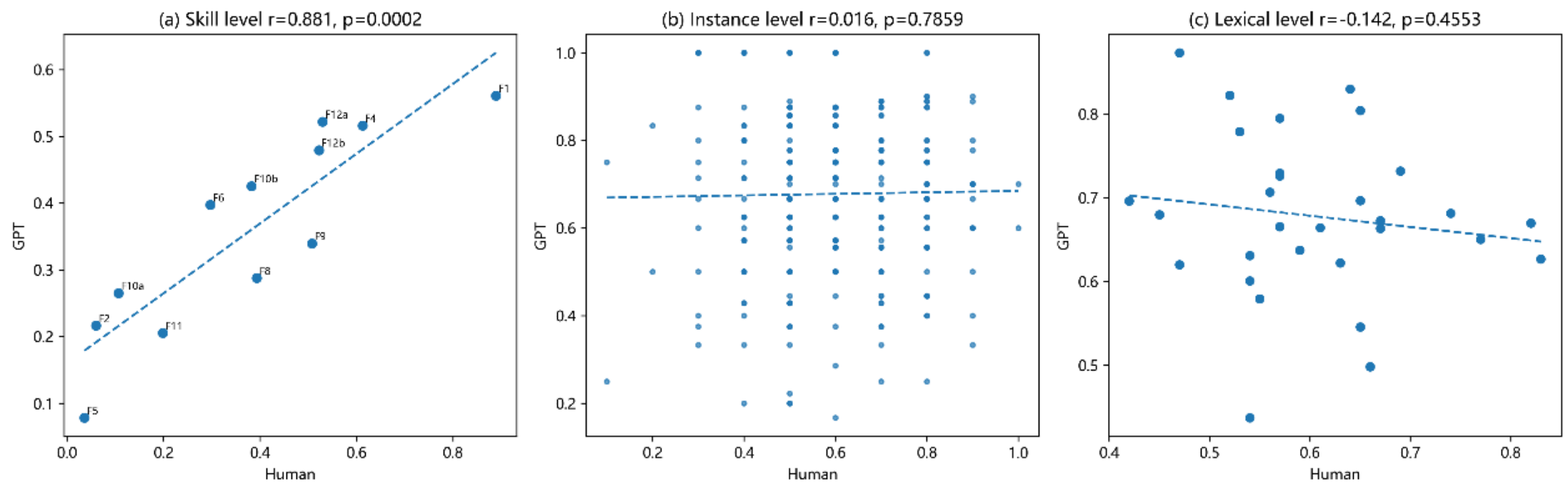

Figure 2. Shared skill difficulty across three levels of analysis

Table 6. Three-level correlations between human and GPT skill difficulty

| level | N | r | p |
|---|---|---|---|
| skill-level shared difficulty | 12 | 0.881 | 0 |
| instance-level independent execution | 300 | 0.016 | 0.786 |
| lexical-level independent execution | 30 | -0.142 | 0.455 |

The most important empirical result of the paper emerges from the three-level analysis. At the skill level, human and GPT difficulty profiles are strongly aligned: the correlation between human and GPT difficulty is $r = 0.8809$. This indicates that the relative difficulty of BP skills is not arbitrary. Skills that are globally difficult for human annotators tend also to be difficult for GPT, while skills that are globally easy for humans tend also to be easier for GPT. In this sense, skill difficulty is shared.

However, this shared structure does not propagate downward. At the instance level, the correlation is only $r = 0.0158$, effectively zero. At the lexical-item level, the correlation is $r = -0.1416$, again indicating no meaningful alignment. These results mean that humans and GPT do not tend to succeed and fail on the same specific instances or the same specific target words, even when they are subject to the same broader skill-difficulty landscape.

This pattern is the empirical basis for the formulation shared taxonomy, independent execution. The "taxonomy" is shared because both human annotators and GPT are constrained by the same skill inventory and broadly track the same difficult and easy skills. The "execution" is independent because, within that shared difficulty topology, the local distribution of errors differs. Humans and GPT agree about where difficulty resides in the schema, but they do not instantiate that difficulty in the same way on individual data points.

Methodologically, this finding matters because it changes how LLM-assisted annotation should be interpreted. If GPT merely copied human error patterns at finer levels, it would add little information beyond what one human annotator already contributes. Instead, the results suggest a more interesting role: GPT participates in the same skill ecology as human annotators, but as an executionally distinct source of annotation evidence.

4.5 GPT's agreement profile occupies a distinct position relative to the human pair

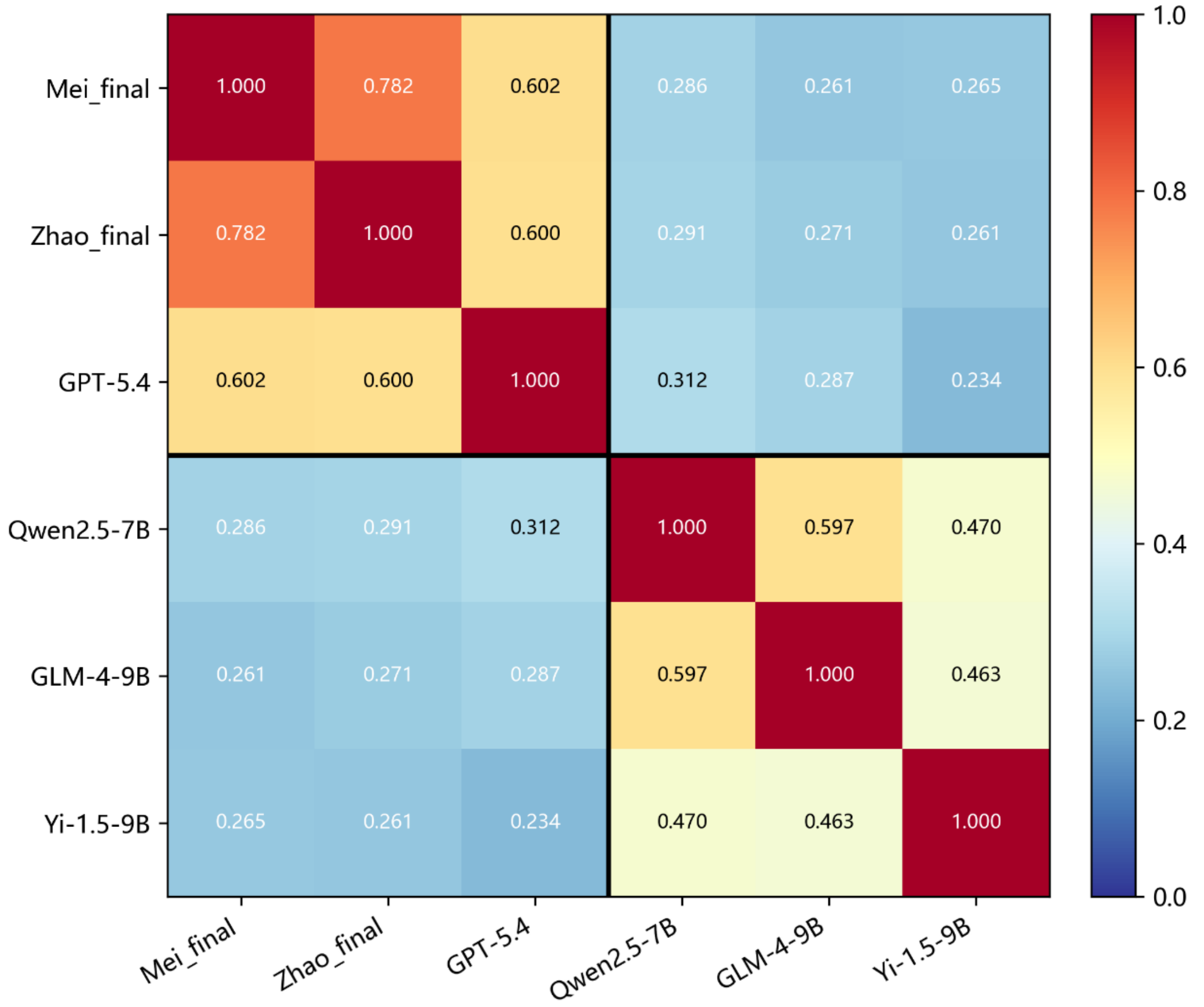

Figure 3. Pairwise agreement matrix among humans, GPT, and open-source models

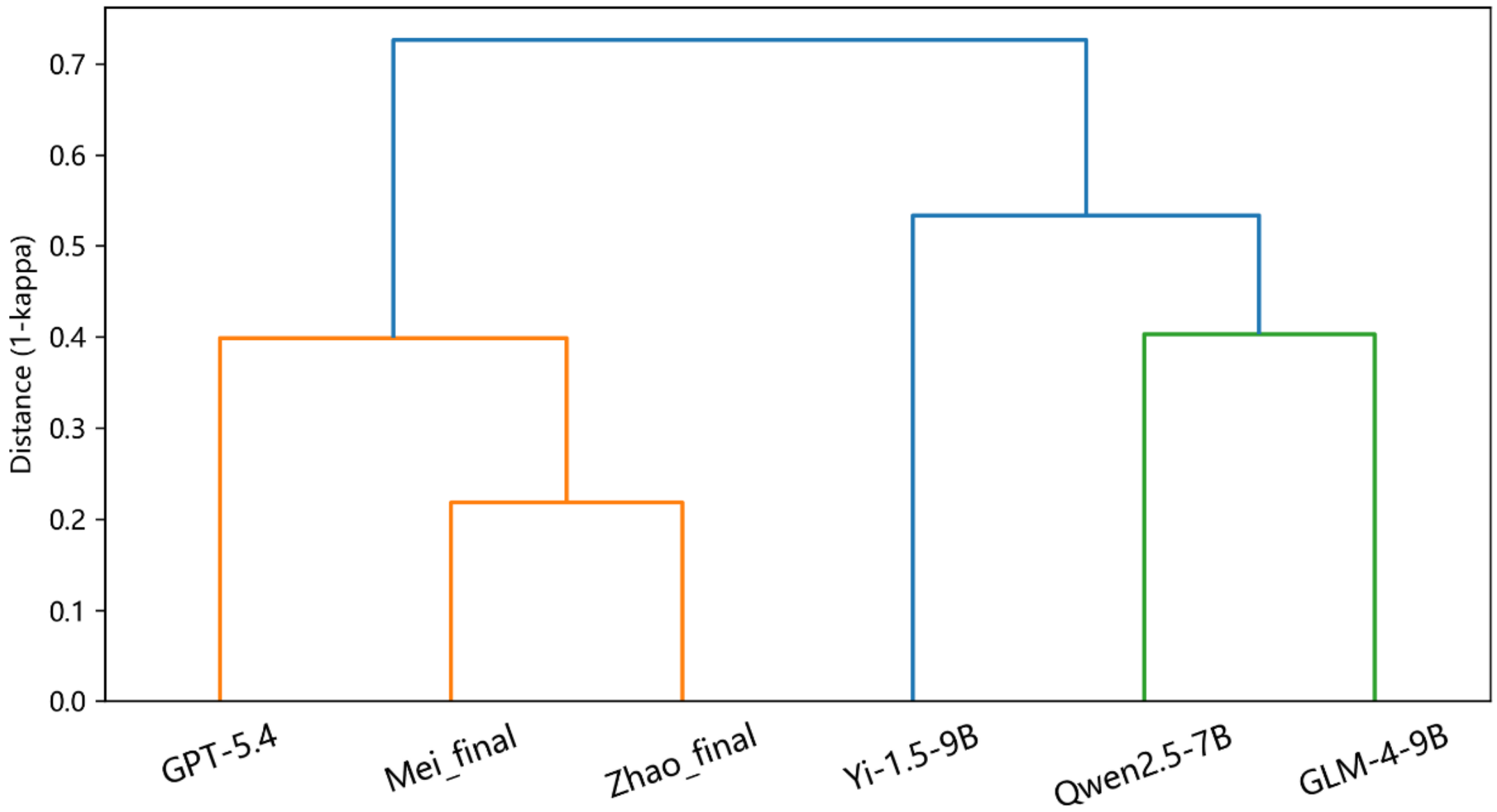

Figure 4. Hierarchical clustering of annotators by pairwise agreement

Table 7. Pairwise kappa matrix among human and model annotators

| | Mei_final | Zhao_final | GPT-5.4 | Qwen2.5-7B | GLM-4-9B | Yi-1.5-9B |
|---|---|---|---|---|---|---|
| Mei_final | 1 | 0.782 | 0.602 | 0.286 | 0.261 | 0.265 |
| Zhao_final | 0.782 | 1 | 0.600 | 0.291 | 0.271 | 0.261 |
| GPT-5.4 | 0.602 | 0.600 | 1 | 0.312 | 0.287 | 0.234 |
| Qwen2.5-7B | 0.286 | 0.291 | 0.312 | 1 | 0.597 | 0.470 |
| GLM-4-9B | 0.261 | 0.271 | 0.287 | 0.597 | 1 | 0.463 |
| Yi-1.5-9B | 0.265 | 0.261 | 0.234 | 0.470 | 0.463 | 1 |

Pairwise agreement analysis clarifies GPT's role further. Human–human agreement is $\kappa = 0.7817$, while GPT's average agreement with the human annotators is $\kappa = 0.6011$, yielding a ratio of 0.7689. Thus, GPT clearly does not match the human pair, but it also performs far above the open-source baseline group and occupies a qualitatively different position in the agreement structure.

The substantive point is not simply that GPT is "close to" human agreement. Rather, GPT appears to function as an independent additional annotation voice: a participant in the same skill space whose outputs are sufficiently aligned with human annotation to be useful, but not so redundant with any individual human annotator that it should be treated as a mere proxy for one of them. This interpretation is supported by the combination of findings already reported: GPT shares human skill difficulty at the feature level, yet shows independent execution patterns at the instance and lexical-item levels.

Under this interpretation, GPT's value lies less in replacing human annotators outright than in increasing the diversity of annotation evidence within the subset of skills that are already operationalizable. This is a weaker claim than "human parity," but it is methodologically more informative. It suggests that the most productive role for GPT is not that of a direct substitute, but that of an independent additional annotator operating within a skill-screened workflow.

### 4.6 Open-source failure is concentrated in schema-to-skill execution

The following analysis of the three open-source models is included as a boundary test for local skill execution under the same skill-file-driven pipeline used for GPT-5.4, not as a primary empirical contribution of the paper. The purpose is to locate where the pipeline breaks when moved from a frontier closed model to 7B–9B open-source models under identical schema-only conditions, and thereby to test whether the skill-feasibility claim behaves differently across model classes. The paper's primary claims about skill feasibility, shared taxonomy and independent execution are not evaluated against the open-source models.

Table 8. Open-source invalid-output rates across BP skills

| model | feature | null_rate | off_schema_rate | invalid_rate |
|---|---|---|---|---|
| Qwen2.5-7B | F1 | 0.360 | 0.000 | 0.360 |
| Qwen2.5-7B | F2 | 0.370 | 0.000 | 0.370 |
| Qwen2.5-7B | F4 | 0.387 | 0.373 | 0.760 |
| Qwen2.5-7B | F8 | 0.340 | 0.000 | 0.340 |
| Qwen2.5-7B | F9 | 0.327 | 0.000 | 0.327 |
| Qwen2.5-7B | F10a | 0.303 | 0.000 | 0.303 |
| Qwen2.5-7B | F10b | 0.303 | 0.000 | 0.303 |
| Qwen2.5-7B | F11 | 0.303 | 0.000 | 0.303 |
| Qwen2.5-7B | F12a | 0.303 | 0.000 | 0.303 |
| Qwen2.5-7B | F12b | 0.310 | 0.000 | 0.310 |
| GLM-4-9B | F1 | 0.307 | 0.000 | 0.307 |
| GLM-4-9B | F2 | 0.320 | 0.000 | 0.320 |
| GLM-4-9B | F4 | 0.310 | 0.447 | 0.757 |
| GLM-4-9B | F8 | 0.323 | 0.000 | 0.323 |
| GLM-4-9B | F9 | 0.303 | 0.000 | 0.303 |
| GLM-4-9B | F10a | 0.303 | 0.000 | 0.303 |
| GLM-4-9B | F10b | 0.303 | 0.000 | 0.303 |
| GLM-4-9B | F11 | 0.303 | 0.000 | 0.303 |
| GLM-4-9B | F12a | 0.303 | 0.000 | 0.303 |
| GLM-4-9B | F12b | 0.310 | 0.000 | 0.310 |
| Yi-1.5-9B | F1 | 0.310 | 0.000 | 0.310 |
| Yi-1.5-9B | F2 | 0.423 | 0.000 | 0.423 |
| Yi-1.5-9B | F4 | 0.747 | 0.020 | 0.767 |
| Yi-1.5-9B | F8 | 0.367 | 0.000 | 0.367 |
| Yi-1.5-9B | F9 | 0.370 | 0.000 | 0.370 |
| Yi-1.5-9B | F10a | 0.550 | 0.000 | 0.550 |
| Yi-1.5-9B | F10b | 0.357 | 0.000 | 0.357 |
| Yi-1.5-9B | F11 | 0.653 | 0.000 | 0.653 |
| Yi-1.5-9B | F12a | 0.383 | 0.000 | 0.383 |
| Yi-1.5-9B | F12b | 0.350 | 0.000 | 0.350 |

The three open-source models do not fail uniformly across the BP skill inventory. Instead, their failures are concentrated in particular skills and often take the form of invalid or off-schema outputs rather than uniformly poor linguistic predictions. The most important example is F4, where off-schema variants such as *AT* and *AD* reveal a mismatch between the schema's expected label inventory and the models' preferred output conventions. This is best interpreted as a schema-to-skill execution failure: the underlying distinction may be partially understood, but the model fails to execute the skill in the constrained label space required by the schema.

This diagnosis matters because it clarifies what kind of failure is being observed. The results do not justify the strongest possible claim that open-source models lack the linguistic capacity for BP annotation in general. What they do support is the narrower and more defensible claim that, under zero-shot schema-only conditions, the tested open-source models are not currently reliable BP skill executors. Some of this unreliability appears to be tied to label-mapping and constrained-execution problems rather than to an absolute ceiling on linguistic knowledge.

At the same time, this qualification should not be overstated. Even if the failure is partly one of schema-to-label execution, that still makes the models practically unsuitable for direct deployment in the current workflow. In other words, the diagnosis is explanatory rather than exculpatory: the models fail for a more specific reason than simple incompetence, but they still fail under the conditions relevant to this study.

### 4.7 A skill-based annotation workflow follows naturally from the results

Taken together, the results support a skill-based BP annotation workflow rather than a task-level automation workflow. The evidence suggests four steps. First, the schema should be used to define the skill inventory explicitly. Second, human pilot annotation should be used to classify skills into directly operable, recoverable, and structurally underspecified categories. Third, GPT should be deployed only on the subset of skills that are sufficiently operationalizable under schema-guided conditions. Fourth, GPT should be integrated not as a full human replacement, but as an independent third skill voice whose outputs contribute to gold construction and adjudication.

Table 9. A skill-based workflow for semi-automatic BP annotation

| step | name | description |
|---|---|---|
| 1 | pilot the skills | Estimate round-1 operability and error modes per skill. |
| 2 | screen skills by operability | Retain directly operable and recoverable skills; separate structural failures. |
| 3 | deploy LLM as a third skill voice | Use LLM as independent skill executor and compare against dual humans. |
| 4 | construct three-way gold | Adjudicate unresolved skills with human-human-LLM triangulation. |

## 5. Discussion

The Discussion develops the methodological implications of the empirical results in eight subsections. Section 5.1 argues for evaluating multi-level annotation at the level of skill feasibility rather than task-level automation. Section 5.2 draws out the consequence of strong skill-level but near-zero instance- and lexical-level correlations: shared difficulty does not imply shared execution. Section 5.3 situates GPT as a third skill voice rather than a human substitute. Section 5.4 distinguishes cognitive-load difficulty from structural difficulty as distinct sources of skill failure with different remedies. Section 5.5 interprets the open-source failures as narrower than global linguistic incapacity while still practically limiting. Section 5.6 generalizes the implications beyond BP annotation to corpus annotation methodology more broadly. Section 5.7 states the limitations, and Section 5.8 offers an overall interpretation.

5.1 From task automation to skill feasibility

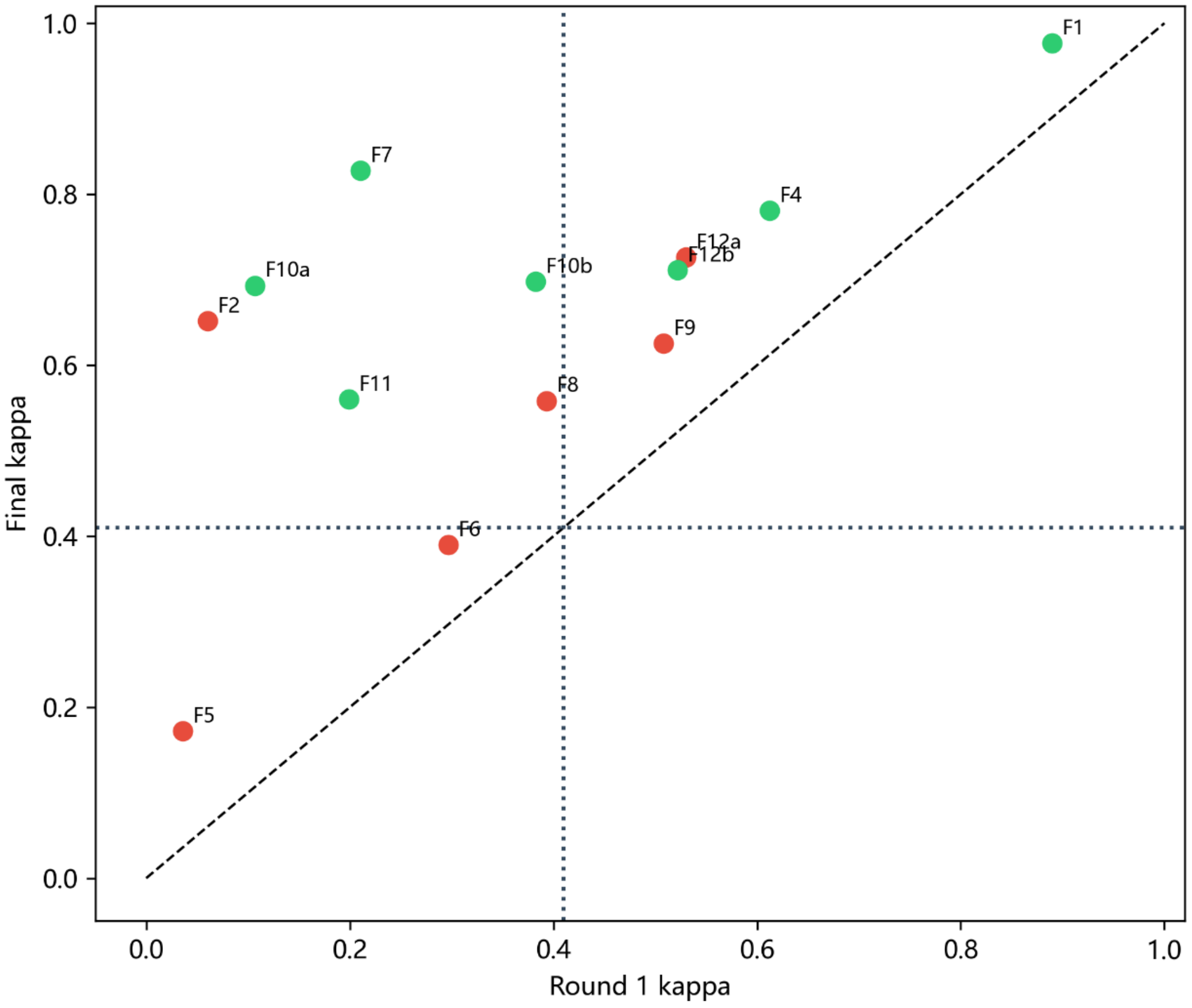

Figure 5. Round-1 versus final skill operability: cognitive-load and structural difficulty

The central theoretical implication of this study is that multi-level BP annotation should not be evaluated as a single automation target. The conventional question—whether "BP annotation" can be automated—presupposes that the task is internally homogeneous enough to admit a single yes-or-no answer. The present results show that this presupposition is untenable. BP annotation is better understood as a structured bundle of annotation skills whose operability varies systematically across the schema. Some skills are directly executable under schema-only conditions, some become reliable only when first-pass cognitive load is reduced, and others remain structurally underspecified. Under these conditions, the relevant methodological unit is not the task as a whole but the individual skill.

This shift from task automation to skill feasibility matters because it changes both what counts as success and what counts as failure. A pooled model score can only be interpreted meaningfully after the skill inventory has been screened for operability. Without that screening, an overall metric would collapse together skills that are directly executable, skills that are merely recoverable, and skills that are fundamentally unstable under the current schema. The present study therefore argues that "automatic annotation" is not a binary property of BP annotation as a whole. It is a conditional property of particular skill units under particular schema-guided execution conditions.

This interpretation also helps explain why the current system architecture matters. Because the annotation pipeline is already structured around externally defined skills—via schema files, prompt templates, runner configuration, and feature-wise execution—skill feasibility is not only an interpretive abstraction but a property of the implemented workflow itself. The paper's methodological claim is therefore not simply that skill framing is a useful metaphor, but that it provides the correct level of analysis for a pipeline that is already skill-file-driven in practice.

### 5.2 Shared difficulty does not imply shared execution

One of the most informative findings of the study is the contrast between strong skill-level correlation and near-zero instance- and lexical-level correlations. At first sight, the feature-level result suggests a reassuring picture: human and GPT difficulty profiles align strongly across the BP skill inventory. But the finer-grained analyses show that this shared structure has clear limits. Humans and GPT agree on which skills are globally difficult, yet they do not systematically fail on the same instances or the same target words. This means that shared difficulty does not collapse into shared error distribution.

This distinction has methodological importance beyond the present dataset. Many LLM-assisted annotation studies report a single accuracy or agreement value and treat that value as evidence of a model's overall human-likeness. The present results suggest that such a summary can be misleading. A model may look human-like in aggregate because it is governed by the same broad difficulty topology as human annotators, while still differing substantially in how it allocates error locally. The right conclusion is therefore not that GPT "thinks like humans," but that GPT and humans are constrained by the same schema-level difficulty structure while remaining executionally distinct.

This is precisely what the formulation shared taxonomy, independent execution is meant to capture. The taxonomy is shared because the schema organizes skills into a broad hierarchy of easier and harder operations that constrains both human and model annotation. Execution is independent because the actual realization of that difficulty is distributed differently across instances and lexical items. In practical terms, this means that GPT contributes non-redundant annotation evidence even when it operates within the same schema-defined difficulty landscape as humans.

### 5.3 GPT as a third skill voice rather than a human substitute

This distinction between shared difficulty and independent execution supports a more precise interpretation of GPT's role in the annotation workflow. The current dominant framing of LLM-assisted annotation often asks whether the model is "good enough to replace a human annotator." That framing is natural but too coarse. In the present study, GPT does not match the human pair, but neither does it behave as a weak approximation of a specific human annotator. Instead, GPT occupies an intermediate but methodologically distinctive position: it participates in the same skill space as the human annotators, achieves substantial agreement on the retained skills, and yet contributes largely independent execution patterns at finer levels.

The term third skill voice is intended to capture this role. GPT is not simply a second-best human. Nor is it a black-box oracle that should be trusted independently of the human pilot process. Rather, it is best treated as an additional voice operating over the same skill inventory, useful precisely because it is neither identical to nor fully substitutable for any one human annotator. This framing is stronger than saying that GPT is "helpful" and weaker than saying that it has reached "human parity," but it is better aligned with the empirical structure of the results.

This interpretation also resolves a practical tension. If GPT were judged only by whether it reaches human-human agreement, its role would appear disappointingly incomplete. If it were judged only by its raw agreement with the gold, its usefulness might be overstated. The third-voice framing avoids both distortions. GPT is useful not because it eliminates the need for human annotation, but because it can participate in a screened subset of skills as an additional, statistically non-redundant source of annotation evidence.

### 5.4 Cognitive-load difficulty and structural difficulty are distinct sources of skill failure

A second major contribution of the paper is the distinction between cognitive-load difficulty and structural difficulty. The two-round schema-only protocol makes this distinction visible in a way that single-round annotation cannot. If a skill performs poorly in Round 1 but becomes reliable in Round 2, the problem is not primarily that the skill is undefined or conceptually incoherent. Rather, the problem is that the first-pass annotation context imposes too much simultaneous burden for the skill to be executed consistently. Such skills are recoverable, and their apparent instability is partly an artifact of multi-skill annotation load.

By contrast, skills that remain unreliable after focused re-annotation are better interpreted as structurally underspecified. These are not merely difficult skills; they are skills whose criteria remain insufficiently operational under the current schema-only regime. The distinction matters because the remedies differ. Cognitive-load difficulty suggests workflow or prompting interventions, such as single-skill prompting, staged annotation, or review-based recovery. Structural difficulty suggests schema revision, feature redesign, or alternative measurement strategies.

This distinction also sharpens how LLM performance should be interpreted. If GPT underperforms on a recoverable skill, its failure may reflect the same kind of first-pass burden that affects human annotators under multi-skill conditions. If GPT underperforms on a structurally underspecified skill, the failure may tell us less about the model than about the limits of the schema as an executable skill definition. In both cases, the proper target of interpretation is the skill and its operationalization, not merely the model.

### 5.5 Open-source failure is narrower than global incapacity, but still practically limiting

The open-source results further illustrate why a skill-based interpretation is useful. A purely task-level reading might conclude that locally deployable 7B–9B models are simply "not ready" for BP annotation. The present results support a narrower and more informative claim. Their failures are strongly concentrated in specific schema-to-label execution problems, especially in F4, where off-schema variants point to a mismatch between the prompted label inventory and the models' preferred output conventions. This is not equivalent to showing that the models lack the underlying linguistic distinction entirely. It shows that, under zero-shot schema-only execution, they are not yet reliable executors of that skill.

This narrower diagnosis is methodologically valuable because it points to concrete interventions. If a failure is due to schema-to-skill execution rather than broad semantic incompetence, then constrained decoding, label-space enforcement, few-shot demonstrations, or feature-specific formatting support may improve performance substantially. That said, the present paper does not test those interventions, and the practical conclusion remains unchanged for current purposes: under the evaluated conditions, the open-source models are not yet deployable BP skill executors on the problematic skills.

In other words, explanation should not be confused with exoneration. The open-source models may be failing for a more localized reason than "lack of language understanding," but they are still failing in a way that blocks direct use in the current workflow. The value of the diagnosis lies in clarifying where future engineering effort should be directed.

### 5.6 Implications for corpus annotation methodology

Although the empirical setting of this paper is BP annotation, the methodological implications are broader. Many corpus annotation projects implicitly assume that if a schema has been written down, then its components are equally executable for both humans and models. The present study suggests otherwise. Schemas may contain skills that are directly operable, skills that are only recoverable under reduced load, and skills that remain structurally unstable. A robust annotation methodology should therefore begin not with full-scale automation, but with a pilot phase that diagnoses the operability profile of the skill inventory.

This has at least three implications. First, pilot annotation should be designed to diagnose skill-level operability, not merely to estimate a single inter-annotator agreement number. Second, LLM-assisted annotation should be treated as selective and skill-specific rather than global and task-wide. Third, gold construction should be seen as a multi-voice process in which LLMs may serve as additional annotation evidence on the subset of skills that have already been shown to be operationalizable. Under this view, human annotation, schema design, and model deployment form an integrated methodological system rather than a sequence of isolated steps.

More broadly, the present results support the view that annotation difficulty is, at least in substantial part, a property of the task definition itself. This does not mean that model architecture and prompt design are unimportant. It means that their effects are filtered through the skill structure imposed by the schema. Skill-based analysis therefore offers a way to connect annotation theory, annotation engineering, and LLM evaluation within a single framework.

### 5.7 Limitations

Several limitations should qualify the claims of this study. First, the empirical domain is narrow: the data concern 30 Chinese metaphorical color-term derivatives and a specific BP skill inventory. Generalization to other lexical domains, languages, or annotation traditions must be demonstrated rather than assumed. Second, the strongest theoretical claim—shared skill difficulty with independent execution—rests on relatively small higher-level samples, especially at the skill and lexical-item levels, and should therefore be interpreted as a strong empirical pattern rather than as a final theoretical law.

Third, the human protocol deliberately reflects schema-only operability, not expert-level BP annotation. This is a strength for the present methodological purpose, since it matches the information conditions given to the model, but it also means that the results should not be read as an estimate of what fully trained BP annotators could achieve under calibration-rich conditions. Fourth, GPT's retained-skill performance is substantial but incomplete; the third-voice framing should not be mistaken for a claim of full human interchangeability. Fifth, the diagnosis of open-source failure as schema-to-skill execution failure is plausible and evidence-based, but remains partly inferential until direct constrained-decoding or few-shot interventions are tested.

Sixth, the three-way skill classification (directly operable, recoverable, structurally underspecified) depends on thresholds on round1_kappa, final_kappa, and the reconciliation rate, stated in Section 4.2. The paper does not report a systematic sensitivity analysis for these thresholds, and small changes in cutoffs could in principle reassign borderline skills between the three tiers—most saliently at the boundary between recoverable and structurally underspecified. The shared-taxonomy, independent-execution pattern at the three levels of analysis does not depend on this classification, but the downstream workflow tiering in Table 9 does, and a sensitivity analysis is left for future work.

Finally, the paper does not yet include a direct with-skill vs without-skill ablation. The study shows that the current annotation system is skill-file-driven and that a skill-based interpretation of the results is both natural and empirically productive. It does not yet prove that skill-file-driven prompting is superior to all alternative prompting regimes. That stronger claim remains for future work.

### 5.8 Overall interpretation

Taken together, the findings support a selective and structured view of automatic annotation. BP annotation is not uniformly automatable, but neither is it uniformly resistant to automation. It decomposes into a skill inventory with different operability profiles, and LLM assistance is most useful when it is aligned to that inventory rather than applied indiscriminately to the task as a whole. GPT succeeds not because it solves BP annotation globally, but because it can execute a screened subset of BP skills with substantial reliability while contributing an independent pattern of local evidence. Open-source models fail not because every BP skill exceeds their capacity, but because the current execution regime does not adequately stabilize schema-to-skill mapping for key features.

## 6. Conclusion

This paper has argued that Behavioral Profile (BP) annotation is better understood as a bundle of annotation skills than as a single task. Using 14 BP features applied to 3,134 concordance lines of 30 Chinese metaphorical color-term derivatives, we examined BP annotation under a skill-file-driven execution framework in which labels, decision rules, examples, and output constraints were externally defined and injected into both human and model annotation workflows. On this basis, the paper asked not whether BP annotation can be automated in general, but which BP skills are operationalizable under schema-guided conditions and how humans and LLMs relate to one another across those skills.

The results support four main conclusions. First, BP annotation is empirically heterogeneous: some skills are directly operable under schema-only conditions, some become reliable only after focused re-annotation, and others remain structurally underspecified. Second, GPT-5.4 can execute a retained subset of BP skills with substantial reliability, but this feasibility is selective rather than global. Third, human and GPT difficulty profiles are strongly aligned at the skill level, but this shared structure collapses at the instance and lexical-item levels. This pattern is captured by the formulation shared taxonomy, independent execution: humans and GPT share the same broad skill-difficulty topology, but distribute their errors differently across specific data points. Fourth, the failures of the open-source models are not uniformly distributed across the skill inventory; instead, they are concentrated in schema-to-skill execution problems, especially in label-space mapping and format compliance.

These findings lead to a more precise interpretation of LLM-assisted annotation. GPT-5.4 is not best understood as a direct human substitute. Rather, within the subset of skills that are sufficiently operationalizable, it is most consistent with the evidence to treat it as something closer to an independent third skill voice: a participant in the same skill space as human annotators, but one whose local execution pattern is not reducible to any one human's errors. This interpretation avoids both overclaiming parity and underestimating utility. GPT is useful not because it removes the need for human annotation entirely, but because it can contribute non-redundant evidence within a screened, skill-aware workflow.

Methodologically, the paper proposes a shift in how multi-level linguistic annotation should be evaluated. Instead of beginning from the question of task-level automation, researchers should begin by reconstructing the schema as a skill inventory, piloting the operability of those skills under human annotation, distinguishing recoverable from structurally unstable skills, and only then deploying LLMs on the subset of skills that the schema can support. Under this view, automation feasibility is not a general property of a model, nor a general property of a task, but a conditional property of specific skill units under explicit execution constraints.

The broader implication is that annotation theory, schema design, and LLM deployment should not be treated as separate methodological domains. They are linked through the operational structure of the skill inventory. A schema that is not executable for humans will not become reliably executable simply because it is handed to a powerful model. Conversely, a model that performs well on screened skills does not thereby validate the task as a whole. The proper target of analysis is the interaction between skill definition, human operability, and model execution. In this sense, the contribution of the present study is as much about annotation methodology as it is about LLM evaluation.

The strongest claim justified by the evidence is therefore a selective one: skill-based automatic annotation is feasible, but only for the subset of BP skills that are operationalizable under schema-guided execution. Future work should test whether this conclusion generalizes across other lexical domains, other annotation traditions, and other model classes, and should directly compare skill-file-driven execution with less structured prompting regimes. But even in its present scope, the study shows that a skill-based perspective yields a more accurate and methodologically useful account of what LLM-assisted annotation can and cannot do.